\pdfoutput=1

\documentclass[11pt]{article}

\usepackage[]{acl}

\usepackage{times}
\usepackage{latexsym}

\usepackage{graphicx}
\usepackage{multirow}
\usepackage{array}
\usepackage{booktabs}

\usepackage[T1]{fontenc}

\usepackage[utf8]{inputenc}

\usepackage{microtype}

\title{Controllable Neural Dialogue Summarization with\\ Personal Named Entity Planning}

\author{Zhengyuan Liu, \ Nancy F. Chen \\
  Institute for Infocomm Research, A*STAR, Singapore \\
  \texttt{\{liu\_zhengyuan,nfychen\}@i2r.a-star.edu.sg}}

\begin{document}
\maketitle

\begin{abstract}
In this paper, we propose a controllable neural generation framework that can flexibly guide dialogue summarization with personal named entity planning. The conditional sequences are modulated to decide what types of information or what perspective to focus on when forming summaries to tackle the under-constrained problem in summarization tasks. This framework supports two types of use cases: (1) Comprehensive Perspective, which is a general-purpose case with no user-preference specified, considering summary points from all conversational interlocutors and all mentioned persons; (2) Focus Perspective, positioning the summary based on a user-specified personal named entity, which could be one of the interlocutors or one of the persons mentioned in the conversation. During training, we exploit occurrence planning of personal named entities and coreference information to improve temporal coherence and to minimize hallucination in neural generation. Experimental results show that our proposed framework generates fluent and factually consistent summaries under various planning controls using both objective metrics and human evaluations.
\end{abstract}

\section{Introduction}
Automatic summarization is the task of compressing a lengthy piece of text to a more concise version while preserving the information of the source content. Extractive approaches select and concatenate salient words, phrases, and sentences from the source to form the summary \cite{lin-bilmes-2011-classEXT,kedzie-2018-contentSelection,liu-etal-2020-conditional}. On the other hand, abstractive approaches generate the summary either from scratch or by paraphrasing important parts of the original text \cite{jing2000cutPaste,gehrmann-2018-bottomUp}. For abstractive summarization to be practically usable, it would require more in-depth comprehension, better generalization, reasoning, and incorporation of real-world knowledge \cite{hovy1999automated,see-2017-pgnet}. While extractive models could suffice for document summarization, abstractive approaches are essential for dialogue summarization to be more easily accessible to users.

\begin{figure}[t]
\centering
\includegraphics[width=7.0cm]{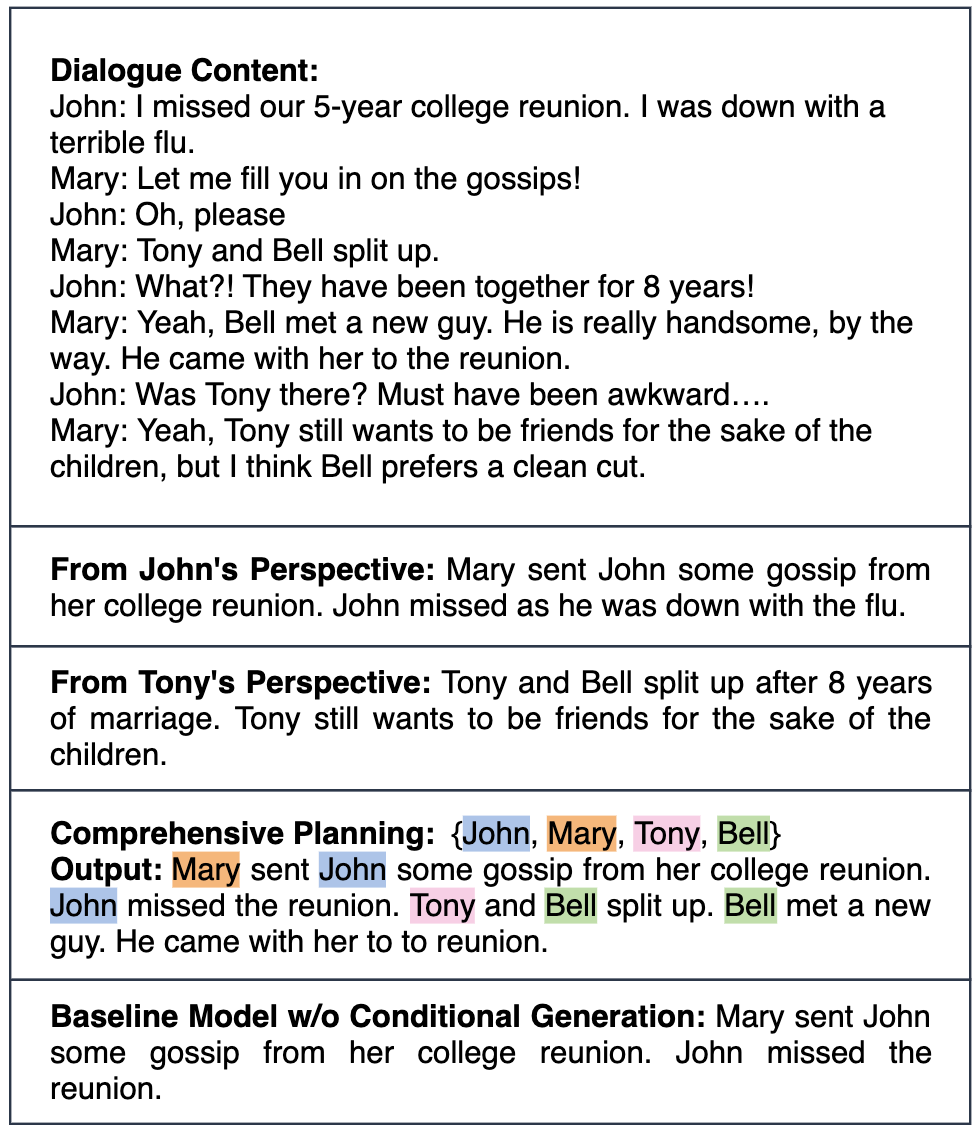}
\caption{Dialogue summary examples generated by personal named entity planning: some examples focus on perspectives from distinct personal named entities (e.g., John, Tony); comprehensive planning includes all personal named entities in the dialogue. Note that the content of the ground-truth summary depends on which personal named entity's perspective the focus is during summary formation.}
\label{sop-cond-dialogue-fig}
\vspace{-0.2cm}
\end{figure}
 
Most benchmarked summarization datasets focus on the news domain, such as NYT \cite{sandhaus-2008-NYT} and CNN/Daily Mail \cite{hermann-2015-cnnDaily} as material for large-scale corpus construction is readily available online. Neural
approaches have achieved favorable improvements in both extractive and abstractive paradigms \cite{paulus-2017-RL-summ,liu-lapata-2019-PreSumm}. Neural dialogue summarization is an emerging research area (e.g., \citet{goo2018-gatedDiaSumm}, \citet{liu-2019-topic}). While the available data collections are much smaller than those for documents \cite{carletta2005ami,gliwa-etal-2019-samsum}, neural models have shown potential to generate fluent sentences via fine-tuning on large scale contextualized language models \cite{chen-yang-2020-MultiView,feng-etal-2021-dialoGPT}.
Unfortunately, most summary generation tasks are constructed in an under-constrained fashion \cite{kryscinski-etal-2019-CriticalSumm}: in their corpus construction process, only one reference summary is annotated. Models trained via supervised learning on such datasets provide general-purpose summaries, but are suboptimal for certain applications and use cases \cite{fan-etal-2018-controllableABS,goodwin2020-NIH-Ctrl}.
For instance, as shown in Figure \ref{sop-cond-dialogue-fig}, a human can write summaries from John or Tony's perspective. However, a neural model with a general summarizing purpose may overlook information that is important to a specific person's perspective. On the other hand, if someone wants to collect as much information from the source content, the summary should be written in a comprehensive manner, taking into consideration all personal named entities. Such needs are not met with models providing only one possible output.

\begin{figure}[t!]
\centering
\includegraphics[width=7cm]{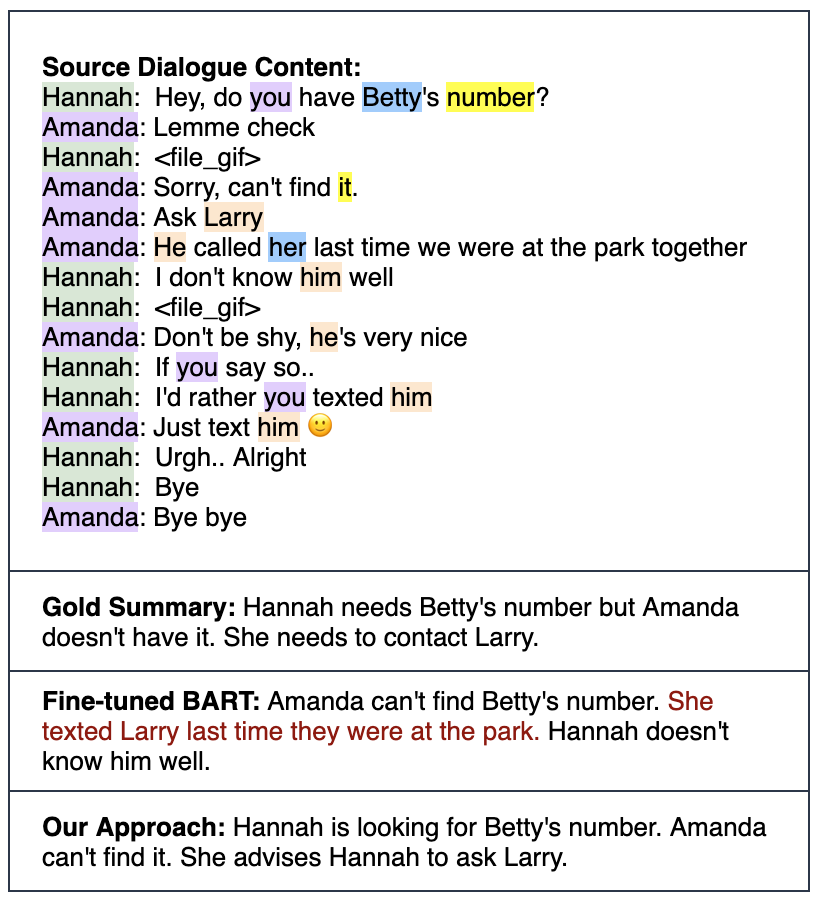}
\caption{One dialogue summarization example: each coreference chain is highlighted with the same color. The generated sentence in red is factually incorrect.}
\label{sop-coref-dialogue-fig}
\vspace{-0.2cm}
\end{figure}

Furthermore, different from passages, human-to-human conversations are a dynamic and interactive flow of information exchange \cite{sacks-1978-simplest}, which are often informal, verbose, and repetitive. 
Since important information is scattered across speakers and dialogue turns, and is often embodied in incomplete sentences. Therefore, generating a fluent summary by utterance extraction is impractical, thus requiring models capable of generating abstractive summaries. However, neural abstractive models often suffer from hallucinations that affect their reliability \cite{zhao-etal-2020-reducing}, involving improper gendered pronouns and misassigned speaker associations \cite{chen-yang-2020-MultiView}.
For example, as shown in Figure \ref{sop-coref-dialogue-fig}, the model makes an incorrect description that \textit{``she texted Larry last time at the park''} (in red). While this sentence achieves a high score in word-overlapping metrics, the semantic meaning it conveys is incorrect: in the context of the generated summary, \textit{she} refers to \textit{Amanda}, yet in reality it is \textit{Larry} that called (not texted) \textit{Betty}. Such factual inconsistency, the inability to adhere to facts from the source, is a prevalent and unsolved problem in neural text generation.

In this work, we introduce a controllable dialogue summarization framework. As the aim of dialogue summaries often focuses on \textit{``who did what''} and the narrative flow usually starts with a subject (often persons), we propose to modulate the generation process with personal named entity plannings. More specifically, as shown in Figure \ref{sop-cond-dialogue-fig}, a set of personal named entities\footnote{A complete named entity set includes personal names, locations, organizations, time expressions, etc.} (in color) are extracted from the source dialogue, and used in a generation model as a conditional signal. We postulate that such conditional anchoring enables the model to support flexible generation. It could be especially useful to address certain demands such as targeting specific client needs for customizing marketing strategies or drilling down customer dissatisfaction at call centers to educate customer agents.
In addition, to improve the quality of conditional generation outputs, we integrate coreference resolution information into the contextual representation by a graph-based neural component to further reduce incorrect reasoning \cite{liu-etal-2021-coreference}.

We conduct extensive experiments on the representative dialogue summarization corpus SAMSum \cite{gliwa-etal-2019-samsum}, which consists of multi-turn dialogues and human-written summaries. Empirical results show that our model can achieve state-of-the-art performance, and is able to generate fluent and accurate summaries with different personal named entity plans. Moreover, factual correctness assessment also shows that the output from our model obtains quality improvement on both automatic measures and human evaluation.

\begin{figure*}[t]
\centering
\includegraphics[width=14cm]{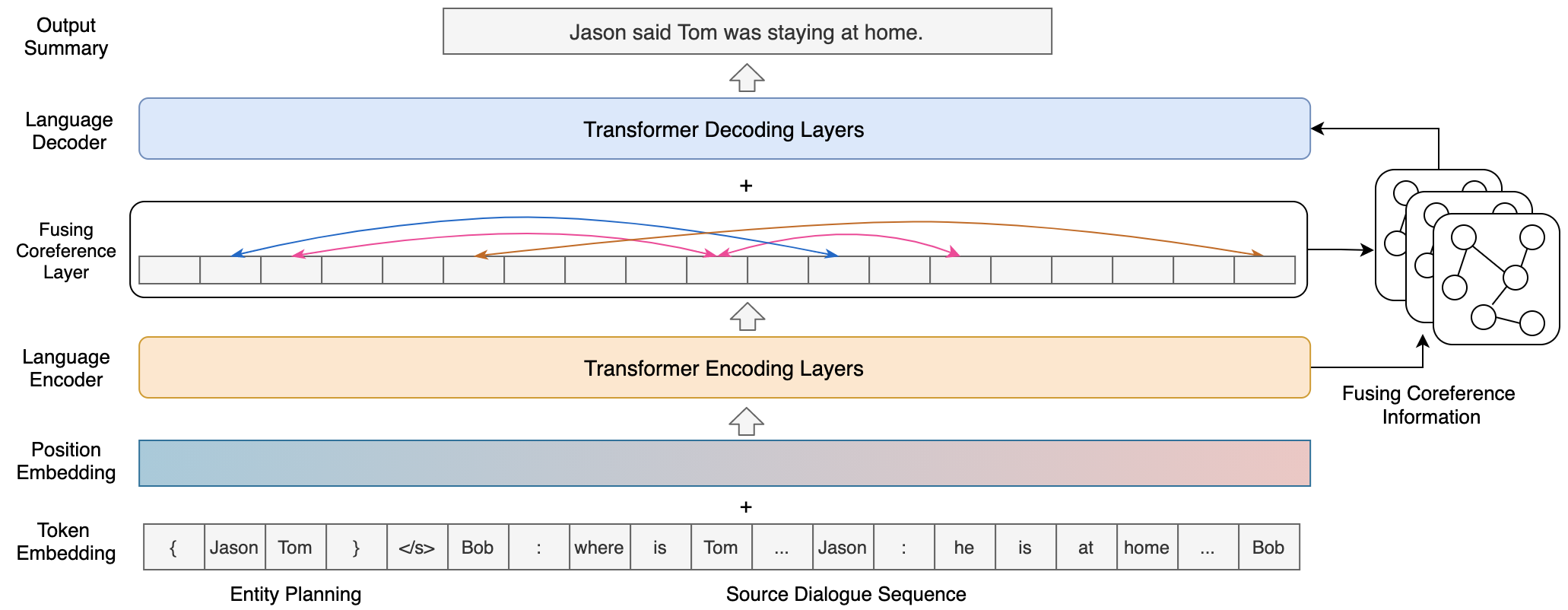}
\caption{Overview of the proposed conditional generation framework with entity planning and coreference integration. Colored lines with arrows in the Fusing Coreference Layer denote the coreference links.}
\label{sop-model-fig}
\end{figure*}

\section{Related Work}
Text summarization has received extensive research attention, and is mainly studied in abstractive and extractive paradigms \cite{gehrmann-2018-bottomUp}. For extractive summarization, non-neural approaches study various linguistic and statistical features via lexical \cite{lexical-summ-1995} and graph-based modeling \cite{erkan-2004-lexrank}. Much progress has been made by recent neural approaches \cite{nallapati2017summarunner, kedzie-2018-contentSelection}.
Compared with extractive methods, abstractive approaches are expected to generate more concise and fluent summaries. While it is a challenging task, with large-scale datasets \cite{hermann-2015-cnnDaily} and sophisticated neural architectures, the performance of abstractive models have achieved substantial improvements in the news domain: sequence-to-sequence models are first introduced by \citet{rush-etal-2015-neuralSumm} for abstractive summarization; pointer-generator network \cite{see-2017-pgnet} elegantly handled out-of-vocabulary issues by copying words directly from the source content; \citet{gehrmann-2018-bottomUp} combines the two paradigms by integrating sentence rewriting into content selection; large-scale pre-trained language models also bring further improvement on summarization performance \cite{liu-lapata-2019-PreSumm, lewis-etal-2020-Bart}.

Recently, neural summarization for conversations has become an emerging research area. Corpora are constructed from meetings \cite{carletta2005ami} or daily chats \cite{gliwa-etal-2019-samsum}. Based on the characteristics of the dialogues, many studies pay attention to utilizing conversational analysis for dialogue summarization, such as leveraging dialogue acts \cite{goo2018-gatedDiaSumm}, multi-modal features \cite{li-etal-2019-keepMeeting}, topic information \cite{liu-2019-topic}, and fine-grained view segmentation with hierarchical modeling \cite{chen-yang-2020-MultiView}.

Controllable text generation introduces auxiliary signals to obtain diverse or task-specific outputs. It has been studied in various domains such as style transferring \cite{shen2017styleTrans} and paraphrasing \cite{iyyer2018adverPara}.
The conditional input can be in the form of pre-defined categorical labels \cite{hu2017towardCtrlText}, latent representations, semantic or syntactic exemplars \cite{gupta2020semanticFrame}, and keyword planning \cite{hua-wang-2020-PairPlan}. Recently, \citet{he2020ctrlsum-SF} and \citet{dou2021gsum} proposed two generic frameworks for length-controllable and question/entity-guided document summarization, and we proposed personal named entity planning upon the characteristics of dialogue summarization.

\section{Controllable Generation with Personal Named Entity Planning}
\label{sec:methodolgy}
In this section, we introduce the proposed conditional generation framework, elaborate on how we construct personal named entity planning, and delineate the steps for training and generation.

\subsection{Task Definition}
Controllable dialogue summarization with personal named entity planning is defined as follows:

\noindent \textbf{Input:} The input consists of two entries: (1) the source content $D$, which is a multi-turn dialogue; (2) a customized conditional sequence $C$, which is the proposed personal named entity planning.

\noindent \textbf{Output:} The output is a natural language sequence $Y$, which represents the summarized information from the source content $D$ with the pre-defined personal named entity plan $C$. Given one instance of $D$, $Y$ can be manifested as various summaries conditioned on different choices of $C$. The output summaries are expected to be fluent and factually correct, covering the indicated entities in the specified conditional signal $C$.

\subsection{Personal Named Entity Planning}
\label{ssec:planning_scheme}

Personal named entities are used to form a planning sequence. 
A customized plan represents what the summary includes, covering specific personal named entities that appear in the dialogue. These named entities \textbf{are not limited to} the speaker roles, but include all persons mentioned in the conversation (e.g., \textit{``Betty''} and \textit{``Larry''} in Figure \ref{sop-coref-dialogue-fig}).

\subsubsection{Training with Occurrence Planning}
\label{ssec:contional_training}
Ground-truth samples for conditional training are built on gold summaries. First, given one dialogue sample and its reference summary, two entity sets are obtained by extracting all personal named entities from the source content and the gold summary respectively. Then, we take the intersection of the two sets, which represent the content coverage of the summary. For instance, given the example in Figure \ref{sop-coref-dialogue-fig}, the intersection is \textit{\{Larry, Amanda, Hannah, Betty\}}.
Next, in order to align the plan with gold summaries written in a certain perspective and narrative flow, we define \textbf{Occurrence Planning}, which reflects the order of personal named entities occurring in the gold summary. To this end, the entity set is re-ordered to \textit{\{Hannah, Betty, Amanda, Larry\}}, and converted to a conditional sequence for training the controllable generation framework.

\subsubsection{Inference: Comprehensive and Focus Planning Summarization Options}
Once the model is trained on personal entity planning, one could customize the input conditional signal as a sequence of personal named entities based on downstream application needs. While our framework supports any combination and order of personal named entities that occurred in the given dialogue, here we focus on two conditional inputs during inference: (1) \textbf{Comprehensive Planning}, which includes all personal named entities in a source dialogue (they are ordered according to the occurrence order in the source) and aims to maximize information coverage. This type of summary supports general purpose use cases. 
(2) \textbf{Focus Planning} only targets one specific personal entity in the dialogue. Focus planning could be viewed as a subset of comprehensive planning and can be useful in more targeted applications.

\subsection{Controllable Neural Generation}
\label{ssec:model_details}
In our framework, a neural sequence-to-sequence network is used for conditional training and generation.
As shown in Figure \ref{sop-model-fig}, the base architecture is a Transformer-based auto-regressive language model, since the Transformer \cite{vaswani-2017-Transformer} is widely adopted in various natural language processing tasks, and shows strong capabilities of contextual modeling \cite{devlin-2019-BERT,lewis-etal-2020-Bart}. The input comprises a source dialogue with $n$ tokens $D=\{w_0, w_1,..., w_n\}$ and a pre-defined personal named entity planning with $m$ tokens $C=\{c_0, c_1,..., c_m\}$.

\noindent\textbf{Encoder:} The encoder consists of a stack of Transformer layers. Each layer has two sub-components: a multi-head layer with self-attention mechanism, and a position-wise feed-forward layer (Equation \ref{eq-attn-head}). A residual connection is employed between each pair of the two sub-components, followed by layer normalization (Equation \ref{eq-attn-residual}).
\begin{equation}
\label{eq-attn-head}
    \widetilde{h}^l=\mathrm{LayerNorm}(h^{l-1}+\mathrm{MHAtt}(h^{l-1}))
\end{equation}
\vspace{-0.4cm}
\begin{equation}
\label{eq-attn-residual}
    h^l=\mathrm{LayerNorm}(\widetilde{h}^l+ \mathrm{FFN}(\widetilde{h}^l))
\end{equation}
where $l$ represents the depth of the stacked layers, and $h^0$ is the embedded input sequence. $\mathrm{MHAtt}$, $\mathrm{FNN}$, $\mathrm{LayerNorm}$ are multi-head attention, feed-forward and layer normalization components, respectively. Moreover, the additional linguistic feature (e.g., coreference information) is added in the encoded representations.

\noindent\textbf{Decoder:} The decoder also consists of a stack of Transformer layers. In addition to the two sub-components in the encoding layers, the decoder inserts another component that performs multi-head attention over hidden representations from the last encoding layer. Then, the decoder generates tokens from left to right in an auto-regressive manner. The architecture and formula details are described in \cite{vaswani-2017-Transformer}.

During training (see Figure \ref{sop-model-fig}), the planning sequence $C$ under \textit{Occurrence Planning} is concatenated with the source dialogue content $D$ as the input with a special token. The segmentation tokens are pre-defined in different Transformer-based models, such as \textit{`[SEP]'} in BERT and \textit{`</s>'} in BART.
The model learns to generate the ground truth $Y=\{y_0, y_1,...,y_t\}$ (where $t$ is the token number) by summarizing the information from the dialogue context conditioned on the planning sequence. The loss of maximizing the log-likelihood on the training data is formulated as:
\begin{equation}
    l(\theta)=-\sum_{t=1}^{T}\mathrm{log}p(y_t|D,C,y_{<t},\theta)
\end{equation}

During inference, we first specify one condition sequence based on the planning schemes described in Section \ref{ssec:planning_scheme}. Specifically, one can assess the model's learning capability by generating summaries guided by \textit{Occurrence Planning}. For simulating the real-world controllable generation scenario, \textit{Comprehensive Planning} and \textit{Focus Planning} can be applied. The model then creates a summary that is based on the specific condition which is coherent with the context of the input conversation.

\section{Improving Factual Correctness}
\label{sec:improve_factual}

While current neural abstractive systems are able to generate fluent summaries, factual inconsistency remains an unsolved problem \cite{zhang-etal-2020-optimizingFC}. Neural models tend to produce statements that are not supported by the source content. These hallucinations are challenging to eradicate in neural modeling due to the implicit nature of learning representations. In document summarization, it has been demonstrated that a certain proportion of abstractive summaries contain hallucinated statements \cite{kryscinski2020evaluating}, as is observed in dialogue summarization \cite{chen-yang-2020-MultiView}. 
Such hallucinations raise concerns about the usefulness and reliability of abstractive summarization, as summaries that perform well in traditional word-overlap metrics may fall short of human evaluation standards \cite{zhao-etal-2020-reducing}.

\subsection{Factual Inconsistency Detection}
\label{ssec:FC_detector}
To evaluate and optimize the summarization quality regarding factual correctness, we first build a model to assess the accuracy of generated statements. Negative samples for classification are built via text manipulation, as is done in prior work \cite{zhao-etal-2020-reducing,kryscinski2020evaluating}. Since we focus on conditional personal named entities in this work, we aim to detect the inconsistency issues of person names between the source content and the generated summaries.

\begin{figure}[t!]
\centering
\includegraphics[width=7.6cm]{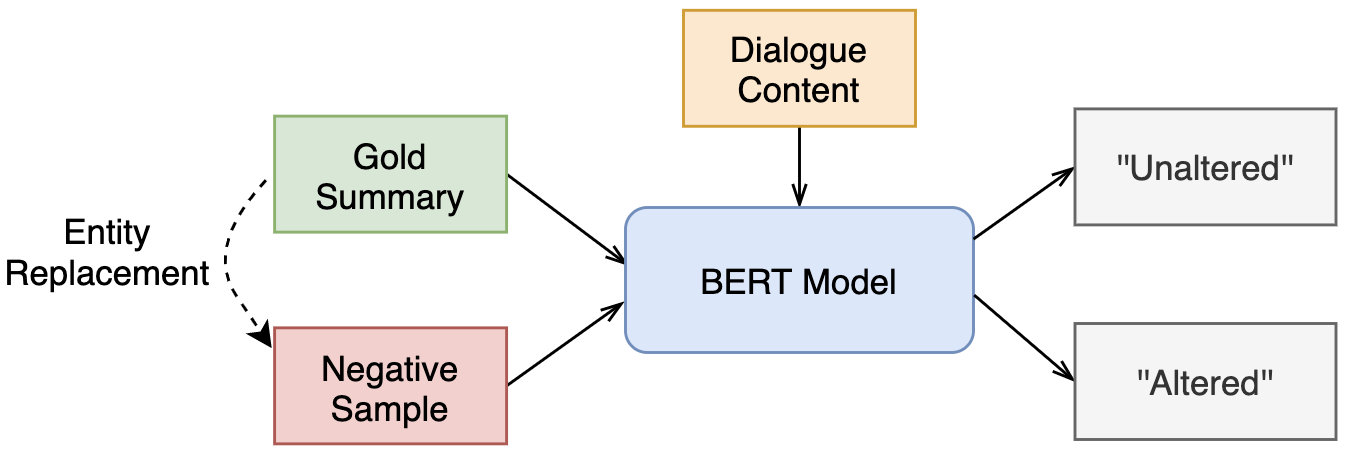}
\caption{Factual inconsistency detection: a binary classification model that determines whether an input summary is altered with named entity replacement.}
\label{sop-fc-detector-fig}
\end{figure}

As shown in Figure \ref{sop-fc-detector-fig}, we construct a binary classifier by reading the dialogue and a summary. The classifier output evaluates if the two input entries are factually consistent.
A reference summary in the original dataset is labeled as \textit{`correct'}.
To generate versions of this summary with label \textit{`incorrect'}, we adopt three strategies to build negative samples: (1) Swapping the positions of where one pair of personal named entities are located in the gold summary with each other. The entities that are connected with word \textit{``and''} and \textit{``or''} in one sentence are excluded; (2) Replacing one name (e.g., \textit{John}) in summaries with another randomly selected name of the same gender (e.g., \textit{Peter}) in the source content; (3) Replacing one name with another from a person name collection built on the training data. With these samples, a \textit{`BERT-base-uncased'} \cite{devlin-2019-BERT} model was fine-tuned to classify whether the summary has been altered. The factual error detector achieved 91\% F1 score on a hold-out validation set.
To identify all personal named entities in both the conversation and the summary, Stanza Named Entity Recognition (NER) tagger \cite{qi2020-stanza} was used.

\subsection{Exploiting Coreference Information}
In conversations, speakers refer to themselves and each other and mention other objects/persons, resulting in various coreference links and chains across dialogue turns and speakers.
We empirically observed that a sizable amount of errors stem from incorrect pronoun assignments in the generation process. Recent language models are also incapable of capturing coreference information without sufficient supervision \cite{dasigi2019quoref}. Thus, we exploit dialogue coreference resolution in a more explicit manner to enhance the model design as in \cite{liu-etal-2021-coreference}.

To this end, we first use the AllenNLP toolkit \cite{Gardner2017AllenNLP} for coreference resolution on the dialogue samples.\footnote{allennlp-public-models/coref-spanbert-large-2021.03.10} With the analyzed coreference mentions and clusters, we build a graph by connecting all nodes in each cluster. Here, we add bi-directional edges between each word/span and its neighboring referring mentions.
Following \cite{liu-etal-2021-coreference}, we incorporate the coreference information into the Transformer-based sequence-to-sequence model. Given a graph with $n$ nodes, we represent the connected structure with an adjacency matrix $A$ where $A_{ij}$ = 1 if node $i$ and node $j$ are connected. For feature integration: (1) to model the linked information with a graph-based method, the multi-layer Graph Convolutional Network (GCN) is applied \cite{kipf2016-GCN}. As shown in Figure \ref{sop-model-fig}, we feed hidden states from the last layer of the language encoder to the graph modeling component, then implicit features are computed and exchanged among tokens in the same coreference cluster, and we add them to the contextualized representation;\footnote{Interested readers can refer to the Appendix for dialogue examples with coreference resolution information.} (2) we also conduct coreference information integration by adding self-attention layers and adopting head manipulation \cite{liu-etal-2021-coreference} which are parameter-efficient, and they can provide the same performance.

\subsection{Data Augmentation via Entity Exchange}
\label{ssec:data_augmentaion}
In addition to the data synthesis strategies in Section \ref{ssec:FC_detector}, we further propose an entity-based data augmentation to robustify the model, reducing incorrect correlations that might be made by the model due to data sparsity or imbalance classes. The augmented data is created by two steps: (1) a personal named entity pair with the same gender attribution is extracted; (2) we exchange them in both source content and reference summary to form new samples. 
For the data used in this paper, each conversation is independent from one another and each interlocutor from a particular dialogue is not an interlocutor in any other dialogue, nor is s/he mentioned in any other dialogue.
Therefore, we postulate that this entity-based augmentation is helpful to reduce unnecessary inductive bias from the training data. In our experiment, the sample number of Data Augmentation (DA) is 4k.

\section{Experimental Results and Analysis}
\label{sec:result_analysis}

\subsection{Dataset Description}
We conduct experiments with the proposed framework on SAMSum \cite{gliwa-etal-2019-samsum}, a dialogue summarization dataset.
It contains multi-turn daily conversations with human-written summaries. The data statistics are shown in Table \ref{table-data-detail}. We retain the original text content of conversations such as cased words, emoticons, and special tokens, and pre-process them using sub-word tokenization \cite{lewis-etal-2020-Bart}. Since the positional embedding of the Transformer-based model can support 1,024 input length, none of the samples are truncated.

\begin{table}[t!]
\linespread{1.1}
\small
\centering
\begin{tabular}{p{4.2cm}p{2.1cm}<{\centering}}
\hline
\textbf{Type} &  \textbf{Number} \\
\hline
\textbf{Training Set} (14732 Samples)  & \\
Mean/Std. of Dialogue Turns &  11.7 (6.45)  \\
Mean/Std. of Dialogue Length &  124.5 (94.2)  \\
Mean/Std. of Summary Length  & 23.44 (12.72) \\
\hline
\textbf{Validation Set} (818 Samples)  & \\
Mean/Std. of Dialogue Turns &  10.83 (6.37)  \\
Mean/Std. of Dialogue Length &  121.6 (94.6)  \\
Mean/Std. of Summary Length  & 23.42 (12.71) \\
\hline
\textbf{Testing Set} (819 Samples) & \\
Mean/Std. of Dialogue Turns &  11.25 (6.35)  \\
Mean/Std. of Dialogue Length &  126.7 (95.7)  \\
Mean/Std. of Summary Length  & 23.12 (12.20) \\
\hline
\end{tabular}
\caption{\label{table-data-detail}Details of the dialogue summarization dataset.}
\vspace{-0.2cm}
\end{table}

\begin{table*}[ht!]
\linespread{1.2}
\small
\begin{center}
\begin{tabular}{p{4.9cm}p{0.65cm}<{\centering}p{0.65cm}<{\centering}p{0.65cm}<{\centering}p{0.65cm}<{\centering}p{0.65cm}<{\centering}p{0.65cm}<{\centering}p{0.65cm}<{\centering}p{0.65cm}<{\centering}p{0.65cm}<{\centering}}
\hline
\multirow{2}{*}{} & \multicolumn{3}{c}{\textbf{ROUGE-1}} & \multicolumn{3}{c}{\textbf{ROUGE-2}} & \multicolumn{3}{c}{\textbf{ROUGE-L}} \\
                        & F       & P       & R       & F       & P       & R       & F       & P       & R       \\
\hline
Pointer Generator*       &    40.1     &   -      &  -       &   15.3      &   -      &   -      &   36.6      &    -     &    -     \\
DynamicConv + GPT-2*     &    41.8     &    -     &    -     &   16.4      &   -      &   -      &    37.6    &   -      &    -     \\
Fast Abs RL Enhanced*    &    42.0     &    -     &    -     &   18.1      &   -      &   -      &   39.2      &    -     &   -      \\
Multi-View BART-Large*   &  49.3  &  51.1  &  52.2  &  25.6  &  26.5  &  27.4  &  47.7  &  49.3  &  49.9    \\
\hline
BART w/o Cond. (Base)       &  50.1  &  56.4  &  49.5  &  25.1  &  28.5  &  24.7  &  47.2  &  51.6  &  46.3    \\
BART w/o Cond. (Large)       &  52.9  &  56.8  &  53.6  &  27.7  &  29.9  &  27.6  &  49.1  &  52.3  &  49.3    \\
\hline
\hline
\multicolumn{3}{l}{Generation with Occurrence Planning}\\
CTRLsum BART-Large (CNN/DM)    &  36.2  &  37.1  &  41.4  &  10.9  &  11.4  &  12.7  &  33.8  &  34.2  &  37.6  \\
CTRLsum BART-Large (Fine-tuned)   &  54.0  &  58.7  &  54.9  &  30.1  &  31.7  &  30.5  &  51.9  &  55.7  &  53.1  \\
\hline
\multicolumn{3}{l}{Generation with Occurrence Planning (ours)}\\
Ctrl-DiaSumm (BART-Base)        &  52.3  &  57.0  &  52.6  &  27.6  &  30.2  &  27.6  &  50.2  &  53.1  &  50.1    \\
Ctrl-DiaSumm+Coref         &  53.5  &  57.7  &  54.3  &  28.9  &  30.9  &  28.7  &  50.4  &  53.2  &  50.5    \\
Ctrl-DiaSumm+Coref+DA         &  53.8  &  58.0  &  55.0  &  29.3  &  31.4  &  29.3  &  51.1  &  53.9  &  51.3    \\
\hline
\multicolumn{3}{l}{Generation with Occurrence Planning (ours)}\\
Ctrl-DiaSumm (BART-Large)        &  54.9  &  56.3  &  57.1  &  30.3  &  31.8  &  32.2  &  52.8  &  54.0  &  54.4    \\
Ctrl-DiaSumm+Coref         &  55.3  &  57.5  &  57.9  &  31.3  &  32.9  &  32.8  &  53.2  &  55.0  &  55.2    \\
Ctrl-DiaSumm+Coref+DA         &  56.0  &  59.8  &  57.6  &  31.7  &  34.4  &  32.2  &  54.1  &  57.8  &  55.3    \\
\hline
\end{tabular}
\end{center}
\caption{\label{output-ROUGE-table} ROUGE scores on the SAMSum test set from baseline models and proposed methods. \textit{Ctrl}, \textit{Coref} and \textit{DA} denote controllable, coreference modeling and data augmentation, respectively. F, P, R are F1 measure, precision, and recall. * denotes the reported results from \cite{chen-yang-2020-MultiView}. \textit{BART w/o Cond.} is the baseline without entity planning conditional training. \textit{CTRLsum} is the generic controllable summarizer proposed in \cite{he2020ctrlsum-SF}, and we further fine-tuned it on the dialogue corpus with our entity planning scheme.}
\end{table*}

\subsection{Model Configurations}
To leverage the large-scale language models which provide semantically-rich contextualized representation to improve downstream tasks such as BERT \cite{devlin-2019-BERT}, we use the implementation of BART that is specially pre-trained for sequence-to-sequence language generation \cite{lewis-etal-2020-Bart},\footnote{https://huggingface.co/facebook/bart-\{base,large\}} to initialize parameters of the Transformer layers in Section \ref{ssec:model_details}, and fine-tune it to boost the performance on our dialogue summarization task.

The number of encoder layers, decoder layers, graph modeling layers, input and hidden dimension are $6/6/2/768$ for the \textit{`BART-Base'} and $12/12/3/1024$ for the \textit{`BART-Large'}, respectively. The learning rate of Transformer layers was set at $3e{-}5$, and that of the graph layers was set at $1e{-}3$. AdamW optimizer \cite{loshchilov2017-adamW} was used with weight decay of $1e{-}3$ and a linear learning rate scheduler. Batch size was set to 8. Drop-out \cite{srivastava2014dropout} of $rate=0.1$ was applied as in the original BART configuration. The backbone parameter size is 139M for the \textit{`BART-Base'} and 406M for for the \textit{`BART-Large'}. For the data augmentation described in Section \ref{ssec:data_augmentaion}, we excluded samples that contain less than two personal named entities in their summaries. Best checkpoints were selected based on validation results of ROUGE-2 value. Tesla A100 with 40G memory was used for training and we used the Pytorch 1.7.1 as the computational framework \cite{paszke2019pytorch}.

\subsection{Quantitative Evaluation}
\label{ssec:automatic_eval_result}
We first conducted two evaluations with automatic metrics to assess the summarizers.
\subsubsection{ROUGE Evaluation}
We adopt ROUGE-1, ROUGE-2, and ROUGE-L, as ROUGE \cite{lin-2004-rouge} is customary in summarization tasks to assess the output quality with gold summaries via counting n-gram overlap. We employ \texttt{Py-rouge} package to evaluate the models following \cite{gliwa-etal-2019-samsum,feng-etal-2021-dialoGPT}.

\begin{table}[t!]
\linespread{1.1}
\centering
\small
\resizebox{\linewidth}{!}
{
\begin{tabular}{p{2.15cm}p{1.3cm}<{\centering}p{1.3cm}<{\centering}p{1.3cm}<{\centering}}
\hline
                & \textbf{Rouge-1 Recall} & \textbf{Rouge-2 Recall} & \textbf{Rouge-L Recall}  \\
\hline
BART w/o Cond.   &  53.6   &  27.6 &  49.3  \\
\hline
\multicolumn{3}{l}{Generation with Comprehensive Planning} \\
CTRLsum*   &  55.7   &  28.2 &  50.8  \\
Ctrl-DiaSumm      &     56.3   &  28.2 &   51.4  \\
Ctrl+Coref     &     58.1   &   28.4  &  52.5 \\
Ctrl+Coref+DA   &     58.4   &    29.1 &  52.9 \\
\hline
\end{tabular}
}
\caption{\label{table-comprehensive-result} ROUGE Recall scores under \textit{Comprehensive Planning}. * \textit{CTRLsum} model is fine-tuned on the dialogue dataset. See complete result table in Appendix.}
\end{table}

\begin{table}[t!]
\linespread{1.1}
\centering
\small
\resizebox{\linewidth}{!}
{
\begin{tabular}{p{2.15cm}p{1.3cm}<{\centering}p{1.3cm}<{\centering}p{1.3cm}<{\centering}}
\hline
                & \textbf{Rouge-1 Precision} & \textbf{Rouge-2 Precision} & \textbf{Rouge-L Precision}  \\
\hline
BART w/o Cond.    &  56.8   & 29.9 &  52.3  \\
\hline
\multicolumn{3}{l}{Generation with Focus Planning} \\
CTRLsum*   &  52.9   &  27.1 &  49.3  \\
Ctrl-DiaSumm      &     52.4   &  27.0 &   49.7  \\
Ctrl+Coref     &     53.1   &   27.2  &  49.9 \\
Ctrl+Coref+DA   &     53.4   &    27.3 &  50.0 \\
\hline
\end{tabular}
}
\caption{\label{table-focus-result} ROUGE Precision scores under \textit{Focus Planning}. * \textit{CTRLsum} model is fine-tuned on the dialogue dataset. See complete result table in Appendix.}
\end{table}

\textbf{Matched Training and Testing Conditions:}
We obtained summaries by conditioning the output generation with the personal named entities in the order they occur in the gold summary (i.e., \textit{Occurrence Planning}). Since \textit{Occurrence Planning} is extracted from the gold summaries, it serves as the upper-bound performance for the proposed conditional generation. As \textit{Comprehensive Planning} and \textit{Focus Planning} are mismatched testing conditions from the training process, we use \textit{Occurrence Planning} to conduct a sanity check to ensure the proposed model performance meets expectations in idealistic scenarios where training and test conditions are matched: Table \ref{output-ROUGE-table} shows that the conditional training in Section \ref{ssec:contional_training} is indeed effective. Moreover, the model with \textit{`BART-Large'} backbone significantly performs better than that of \textit{`BART-Base'}, thus we use it for the following generation evaluations. We also select a generic controllable model \textit{CTRLsum} \cite{he2020ctrlsum-SF} for comparison. We observed that the original \textit{CTRLsum} trained on the news domain cannot generalize well on the dialogue corpus, and the performance can benefit from further fine-tuning.

\noindent\textbf{Mismatched Training \& Testing Conditions:}
As \textit{Comprehensive Planning} covers the maximum number of personal entities in the dialogue, recall (a sensitivity measure) is more suitable in assessing its performance. Similarly, as \textit{Focus Planning} only concerns a specific personal entity, precision (a specificity measure) is adopted. For evaluating \textit{Focus Planning}, we randomly selected one speaker entity from each dialogue as condition input. While the aim of conditional summary generation (be it \textit{Comprehensive Planning} or \textit{Focus Planning}) is not to generate a summary that emulates the gold summary, we nonetheless provide comparison results with the unconditional baseline \textit{`BART w/o Cond.'} for analysis purposes (see Appendix for complete ROUGE results and the generated summary examples).
Results in Table \ref{table-comprehensive-result} - \ref{table-focus-result} suggest: (1) Increasing the information coverage on personal named entities in dialogues improves general-purpose dialogue summarization; (2) Obtaining a reasonably accurate summary focused on a specified personal named entity is feasible; (3) Integrating coreference information and data augmentation improve performance consistently.

\begin{table}[t!]
\linespread{1.1}
\centering
\small
\begin{tabular}{p{3.6cm}p{1.6cm}<{\centering}}
\hline
                & \textbf{Accuracy} \\

\hline
BART w/o Cond. &     77.4       \\
\hline

\multicolumn{2}{l}{\textbf{Occurrence Planning}} 
  \\
CTRLsum (fine-tuned)  &  80.1  \\
Ctrl-DiaSumm          &     79.9    \\
Ctrl+Coref     &    81.5     \\
Ctrl+Coref+DA &    82.8     \\
\hline
\textbf{Comprehensive Planning} &  \\
CTRLsum (fine-tuned)  &  79.5  \\
Ctrl-DiaSumm          &    79.0     \\
Ctrl+Coref     &    80.8     \\
Ctrl+Coref+DA &    81.9     \\
\hline
\textbf{Focus Planning} &  \\
CTRLsum (fine-tuned)  &  74.9  \\
Ctrl-DiaSumm          &    74.3     \\
Ctrl+Coref     &    75.5     \\
Ctrl+Coref+DA &    76.2     \\
\hline
\end{tabular}
\caption{\label{table-FC-accuracy} Automatic factual correctness evaluation on samples from baselines and our models.}
\end{table}

\subsubsection{Factual Correctness Evaluation}
\label{sssec:FC_eval}
We applied the factual consistency classifier built in Section \ref{ssec:FC_detector} to assess the generated summaries using the accuracy metric (the proportion of samples that are predicted as true).
As shown in Table \ref{table-FC-accuracy}, explicitly incorporating coreference information improves the accuracy of generated summaries guided with all conditional plannings, and data augmentation brings further improvements. Results of \textit{Comprehensive Planning} is close to the upper-bound of \textit{Occurrence Planning}. The difference is potentially due to the relatively longer generated summaries and the use of more novel words. Specifically, we observed that the novel word rate \cite{see-2017-pgnet} of \textit{Ctrl+Coref} under \textit{Occurrence} and  \textit{Comprehensive} plannings are 0.28 and 0.33 respectively.
The overall accuracy under \textit{Focus Planning} is relatively lower, which is not unexpected, as more paraphrasing is needed for summarizing from a specified personal entity's perspective. Moreover, the fine-tuned \textit{CTRLsum} performs similarly to the \textit{Ctrl-DiaSumm} model, since both of them use \textit{`BART-Large'} as the language backbone. However, here we did not pre-trained our models on out-of-domain summarization data.

\begin{table}[t!]
\linespread{1.1}
\centering
\small
\begin{tabular}{p{2.8cm}cc}
\hline
                & \textbf{Consistency} & \textbf{Informative} \\

\hline
BART w/o Cond. &     0.71    &  0.70 \\
\hline
\multicolumn{3}{l}{\textbf{Occurrence Planning}} \\
Ctrl-DiaSumm          &     0.78   & 0.79 \\
Ctrl+Coref+DA &    0.79   &  0.81 \\
\hline
\multicolumn{3}{l}{\textbf{Comprehensive Planning}} \\
Ctrl-DiaSumm          &    0.74    & 0.83 \\
Ctrl+Coref+DA &    0.78   &  0.85 \\
\hline
\multicolumn{3}{l}{\textbf{Focus Planning}} \\
Ctrl-DiaSumm          &    0.68    & 0.70 \\
Ctrl+Coref+DA &    0.75   &  0.77 \\
\hline
\end{tabular}
\caption{\label{table-human-quality} Quality scoring on generated samples from models. Scores are normalized in the range of $[0,1.0]$.}
\end{table}

\begin{table*}[t!]
\linespread{1.1}
\small
\centering
\begin{tabular}{lp{2.2cm}<{\centering}p{2.0cm}<{\centering}p{2.3cm}<{\centering}p{2.0cm}<{\centering}p{2.3cm}<{\centering}}
\hline
   &   & \multicolumn{2}{c}{\textbf{Comprehensive Planning}}   & \multicolumn{2}{c}{\textbf{Focus Planning}} \\
   & BART w/o Cond.  & Ctrl-DiaSumm  & Ctrl+Coref+DA   & Ctrl-DiaSumm  & Ctrl+Coref+DA \\
\hline
Average/Std. Length &     21.3 (12.2)     &    26.52  (13.3)  &   27.05 (13.9)  &   15.44  (8.5)  &     15.87 (8.7)  \\
\hline
Missing Information &     17      &    6   &  4 [33\% $\downarrow$]  &  16      &     12  [25\% $\downarrow$]  \\
Wrong References    &    11       &    11   &  8  [27\% $\downarrow$] &   14    &    11 [21\% $\downarrow$]  \\
Incorrect Reasoning &     10       &   12     &  9  [25\% $\downarrow$]  &  13   &   10  [23\% $\downarrow$]  \\
Improper Gender     &     2       &   2   &   1 [50\% $\downarrow$] &     5   &  3 [40\% $\downarrow$] \\
\hline
\end{tabular}
\caption{\label{table-error-analysis} Error analysis on 50 samples. Values in round brackets denote standard deviations of length. Numbers are counted if one error is labeled in generated summaries. Values in square brackets denote the relative decrease.}
\end{table*}

\subsection{Human Evaluation}
\subsubsection{Quality Scoring}
We randomly sampled 50 dialogues with generated summaries for two linguistic evaluators to conduct quality scoring \cite{paulus-2017-RL-summ}. Since abstractive models fine-tuned on contextualized language backbones are able to generate fluent sentences \cite{lewis-etal-2020-Bart,chen-yang-2020-MultiView}, we excluded \textit{fluency} in the scoring criteria. Instead, \textit{factual consistency} and \textit{informativeness} were used to measure how accurate and comprehensive the extracted information is according to the source content. Summaries were scored of $[-1, 0, 1]$, where $-1$ means a summary was not factually consistent or failed to extract relevant information, $1$ means it could be regarded as a human-written output, and $0$ means it extracted some relevant information or made minor mistakes. We averaged the normalized scores from evaluators. As shown in Table \ref{table-human-quality}, \textit{Comprehensive Planning} obtains slightly lower scores in \textit{consistency} (related to ROUGE precision score) than the training scheme \textit{Occurrence Planning}, but it achieves higher \textit{informativeness} scores, which is consistent with the improvement on ROUGE recall scores in Table \ref{table-comprehensive-result}. Moreover, the proposed model (\textit{Ctrl+Coref+DA}) outperforms base model significantly under \textit{Focus Planning}.

\subsubsection{Error Analysis}
Similar to previous work \cite{chen-yang-2020-MultiView}, we conducted error analysis by checking the following 4 error types: (1) Missing information: content mentioned in references is missing in generated summaries; (2) Wrong references: generated summaries contain information that is not faithful to the original dialogue, or associate actions with wrong named entities; (3) Incorrect reasoning: the model learned incorrect associations leading to wrong conclusions in the generated summary; (4) Improper gendered pronouns.
Linguistic analysts were given 50 dialogues randomly chosen from the test set and their corresponding summaries from baselines and our models. They were asked to read the dialogue content and summaries and judge if the 4 error types occurred. For each evaluator, the sequence of presentation was randomized differently. As shown in Table \ref{table-error-analysis}, summaries under both planning schemes make significantly fewer errors at all fronts. Under \textit{Comprehensive Planning}, models with conference information and data augmentation (\textit{Ctrl+Coref+DA}) outperform the base model especially in consistency-related classes. Under \textit{Focus Planning}, both models produce more factual incorrectness due to more paraphrasing from various personal perspectives, this matches the result from automatic factual consistency evaluation in Section \ref{sssec:FC_eval}, and the \textit{Ctrl+Coref+DA} model achieves significant quality improvement.

\section{Conclusion}
In this work, we proposed a controllable neural framework for abstractive dialogue summarization. In particular, a set of personal named entities were used to condition summary generation. This framework could efficiently tailor to different user preferences and application needs, via modulating entity planning. Moreover, the experimental results demonstrated that the abstractive model could benefit from explicitly integrating coreference resolution information, achieving better performance on factual consistency and standard metrics of word-overlap with gold summaries.

\section*{Acknowledgments}
This research was supported by funding from the Institute for Infocomm Research (I2R) under A*STAR ARES, Singapore. We thank Ai Ti Aw for the insightful discussions. We also thank the anonymous reviewers for their precious feedback to help improve and extend this piece of work.

\bibliography{acl_latex}
\bibliographystyle{acl_natbib}


\appendix

\section{Appendix}
\label{sec:appendix}

\begin{table*}[ht!]
\linespread{1.2}
\small
\begin{center}
\begin{tabular}{p{5cm}p{0.7cm}<{\centering}p{0.7cm}<{\centering}p{0.7cm}<{\centering}p{0.7cm}<{\centering}p{0.7cm}<{\centering}p{0.7cm}<{\centering}p{0.7cm}<{\centering}p{0.7cm}<{\centering}p{0.7cm}<{\centering}}
\hline
\multirow{2}{*}{} & \multicolumn{3}{c}{\textbf{ROUGE-1}} & \multicolumn{3}{c}{\textbf{ROUGE-2}} & \multicolumn{3}{c}{\textbf{ROUGE-L}} \\
                        & F       & P       & R       & F       & P       & R       & F       & P       & R       \\
\hline
\multicolumn{3}{l}{Generation with Comprehensive Planning on Subset-A}\\
CtrlSum (fine-tuned) \cite{he2020ctrlsum-SF}        &  54.1  &  56.1  &  55.3  &  27.1  &  29.3  &  27.8  &  48.7  &  50.8  &  48.5    \\
Ctrl-DiaSumm          &  53.7  &  54.3  &  55.9  &  25.7  &  26.9  &  27.9  &  48.6  &  49.9  &  49.4    \\
Ctrl+Coref         &  53.9  &  55.7  &  56.9  &  27.1  &  28.1  &  28.0  &  49.0  &  50.1  &  51.0    \\
Ctrl+Coref+DA         &  54.6  &  56.5  &  57.4  &  27.6  &  29.1  &  28.6  &  49.7  &  51.0  &  51.5    \\
\hline
\multicolumn{3}{l}{Generation with Comprehensive Planning on Subset-B}\\
CtrlSum (fine-tuned) \cite{he2020ctrlsum-SF}   &  46.7  &  47.2  &  56.1  &  24.0  &  23.1  &  28.3  &  44.6  &  43.7  &  51.5    \\
Ctrl-DiaSumm          &  47.3  &  43.8  &  57.6  &  23.5  &  21.7  &  29.1  &  43.3  &  40.8  &  51.7    \\
Ctrl+Coref         &  47.9  &  44.3  &  59.1  &  23.7  &  22.1  &  29.4  &  44.0  &  41.2  &  53.2    \\
Ctrl+Coref+DA         &  48.4  &  44.7  &  59.7  &  24.4  &  22.4  &  30.8  &  45.2  &  41.7  &  54.0    \\
\hline
\end{tabular}
\end{center}
\caption{\label{appendix-FOCUS-result} ROUGE scores on summaries under the \textit{Comprehensive Planning}. \textit{Ctrl}, \textit{Coref} and \textit{DA} denote controllable, coreference modeling and data augmentation, respectively. F, P, R are F1 measure, precision, and recall. For fair comparison with ground-truth summaries, we split the test set into two subsets: Subset-A (461 of 819 test samples) contains the samples that personal entity set extracted from gold summaries and that of \textit{Comprehensive Planning} is the same, and the rest 358 samples are included in Subset-B.}
\vspace{-0.5cm}
\end{table*}

\begin{table*}[ht!]
\linespread{1.2}
\small
\begin{center}
\begin{tabular}{p{5cm}p{0.7cm}<{\centering}p{0.7cm}<{\centering}p{0.7cm}<{\centering}p{0.7cm}<{\centering}p{0.7cm}<{\centering}p{0.7cm}<{\centering}p{0.7cm}<{\centering}p{0.7cm}<{\centering}p{0.7cm}<{\centering}}
\hline
\multirow{2}{*}{} & \multicolumn{3}{c}{\textbf{ROUGE-1}} & \multicolumn{3}{c}{\textbf{ROUGE-2}} & \multicolumn{3}{c}{\textbf{ROUGE-L}} \\
                        & F       & P       & R       & F       & P       & R       & F       & P       & R       \\
\hline
\multicolumn{3}{l}{Generation with Focus Planning}\\
CtrlSum (fine-tuned) \cite{he2020ctrlsum-SF}   &  47.0  &  52.9  &  45.7  &  23.3  &  27.1  &  23.1  &  44.5  &  49.3  &  43.7    \\
Ctrl-DiaSumm          &  47.0  &  52.4  &  46.8  &  23.0  &  27.0  &  22.8  &  44.8  &  49.7  &  44.1    \\
Ctrl+Coref         &  47.1  &  53.1  &  47.0  &  23.4  &  27.2  &  23.3  &  45.1  &  49.9  &  44.6    \\
Ctrl+Coref+DA         &  47.4  &  53.4 &  47.9  &  23.8  &  27.3  &  23.9  &  45.3  &  50.0  &  45.0    \\
\hline
\end{tabular}
\end{center}
\caption{\label{appendix-EXH-result} ROUGE scores on summaries under the \textit{Focus Planning}. \textit{Ctrl}, \textit{Coref} and \textit{DA} denote controllable, coreference modeling and data augmentation, respectively. F, P, R are F1 measure, precision, and recall. For the \textit{Focus Planning}, we randomly selected one speaker entity from each dialogue as condition input. Worth-mentioned that the average length of generation under \textit{Focus Planning} is smaller than that of \textit{Comprehensive Planning}, resulting in some decrease of recall performance. Moreover, more paraphrasing is needed for generated difference summaries from different personal perspectives, as the examples shown in Table \ref{appendix-focus-examples}.}
\vspace{-0.5cm}
\end{table*}

\begin{table*}[ht!]
\linespread{1.1}
\small
\centering
\begin{tabular}{lp{2.2cm}<{\centering}p{2.0cm}<{\centering}p{2.3cm}<{\centering}p{2.0cm}<{\centering}p{2.3cm}<{\centering}}
\hline
   &   & \multicolumn{2}{c}{\textbf{Occurrence Planning}}   & \multicolumn{2}{c}{\textbf{Comprehensive Planning}} \\
   & BART w/o Cond.  & Ctrl-DiaSumm  & Ctrl+Coref+DA   & Ctrl-DiaSumm  & Ctrl+Coref+DA \\
\hline
Average/Std. Length &     21.3 (12.2)     &    20.96 (9.75)  &  19.82 (9.47)  &   26.52  (13.3)  &     27.05 (13.9)  \\
\hline
Missing Information &     17       &    6   &  4 [33\% $\downarrow$]  &   6    &     4  [33\% $\downarrow$]  \\
Wrong References    &    11        &    7   &  6   [14\% $\downarrow$] &   11    &    8   [27\% $\downarrow$] \\
Incorrect Reasoning &     10       &  9     &  8  [11\% $\downarrow$]  &   12    &   9  [25\% $\downarrow$] \\
Improper Gender     &     2       &   1    &   1  [0\% $\downarrow$] &    2   &    1   [50\% $\downarrow$] \\
\hline
\end{tabular}
\caption{\label{table-occur-error-analysis} Error analysis on 50 samples from the baseline and our models. Here we compare the \textit{Comprehensive Planning} with the training scheme \textit{Occurrence Planning}. Values in round brackets are the standard deviation of summary length. Numbers are counted if there is an error labeled in the generated summary. Values in square brackets are the relative decrease.}
\vspace{-0.7cm}
\end{table*}

\begin{table*}[ht]
\linespread{1.1}
    \centering
    \small
    \begin{tabular}{m{.52\textwidth}| m{.2\textwidth}| m{.2\textwidth}}
    \hline
         \textbf{Conversation} & \textit{\textbf{Reference Summary}} & \textit{\textbf{Focus Planning}}\\
         \hline
          (\romannumeral1)  <Natalie>: Well well weeeeeell, I see somethings going on here at last. <Martin>: (Y). Adam: any confirmed data? <Anna>: Hello everyone!!! Id love to invite everybody to my bday. I would be extremaly happy if you could come 6th of November at 19:30. <Martin>: (smile)] <Margot>: (smile) <Mia>: (Y) 
          & Anna organises a birthday's party on the 6th of November at 19:30.
          & \textcolor{blue}{\textbf{Adam}} will come to Anna's birthday party on 6th November at 19:30.
          \newline
          ------------------------------
          \newline
          \textcolor{blue}{\textbf{Anna}} invites everyone to her birthday on 6th November at 19:30. \\
         \hline
          (\romannumeral2) <Anne>: You were right, he was lying to me :/ <Irene>: Oh no, what happened? <Jane>: who? that Mark guy? <Anne>: yeah, he told me he's 30, today I saw his passport - he's 40 <Irene>: You sure it's so important? <Anne>: he lied to me Irene.
          & Mark lied to Anne about his age. Mark is 40.
          & \textcolor{blue}{\textbf{Anne}} saw a man's passport today. He's 40.
          \newline
          ------------------------------
          \newline
          \textcolor{blue}{\textbf{Jane}}'s friend lied to her about him being 30 years old. 
          \\
          \hline
          (\romannumeral3) <Josh>: Stephen, I think you've accidentaly taken my notebook home <Stephen>: wait lemme check. <Stephen>: nope, I don't see it anywhere <Jack>: oh xxx, I've got it xDDD I don't even know why. <Josh>: xDDD ok, no problem, cool I know where it is. <Jack>: I'll bring it tomorrow.
          & Josh thinks Stephen accidentally took his notebook. Jack has it and will bring it tomorrow. 
          &\textcolor{blue}{\textbf{Jack}} has taken Stephen's notebook and will bring it tomorrow.
          \newline
          ------------------------------
          \newline
          \textcolor{blue}{\textbf{Stephen}} has left his notebook at home. He can't find it.    
            \\
          \hline
          (\romannumeral4) <George>: What have you gotten for Christmas? <Jacob>: I got a punchbag. <Jenny>: I got training shoes. <George>: Sporty team :P <Jenny>: What did you get? <George>: A cooking pot :-) <Jacob>: Your wife wants you to help her in the kitchen? <George>: It's me who is normally cooking. <George>: I really like it :P <George>: Jenny gave me this pot, it's amazing and has life long guarantee. <Jacob>: Cool <Jenny>: I wish my Michael was a better cook. <Jenny>: I think it's really sexy when a guy can cook well. 
          & Jacob, Jenny and George are telling each other what they have gotten for Christmas. 
          &\textcolor{blue}{\textbf{George}} got a cooking pot for Christmas. His wife wants him to help her in the kitchen.
          \newline
          ------------------------------
          \newline
          \textcolor{blue}{\textbf{Jenny}} got a sports bag for Christmas, a cooking pot and training shoes.    
            \\
          \hline
    \end{tabular}
    \caption{Examples of generated summaries with \textit{Focus Planning}. Speaker roles are bracketed, and the focused personal named entity is highlighted.}
    \label{appendix-focus-examples}
\vspace{-0.2cm}
\end{table*}

\begin{figure*}[hb]
\centering
\includegraphics[width=15cm]{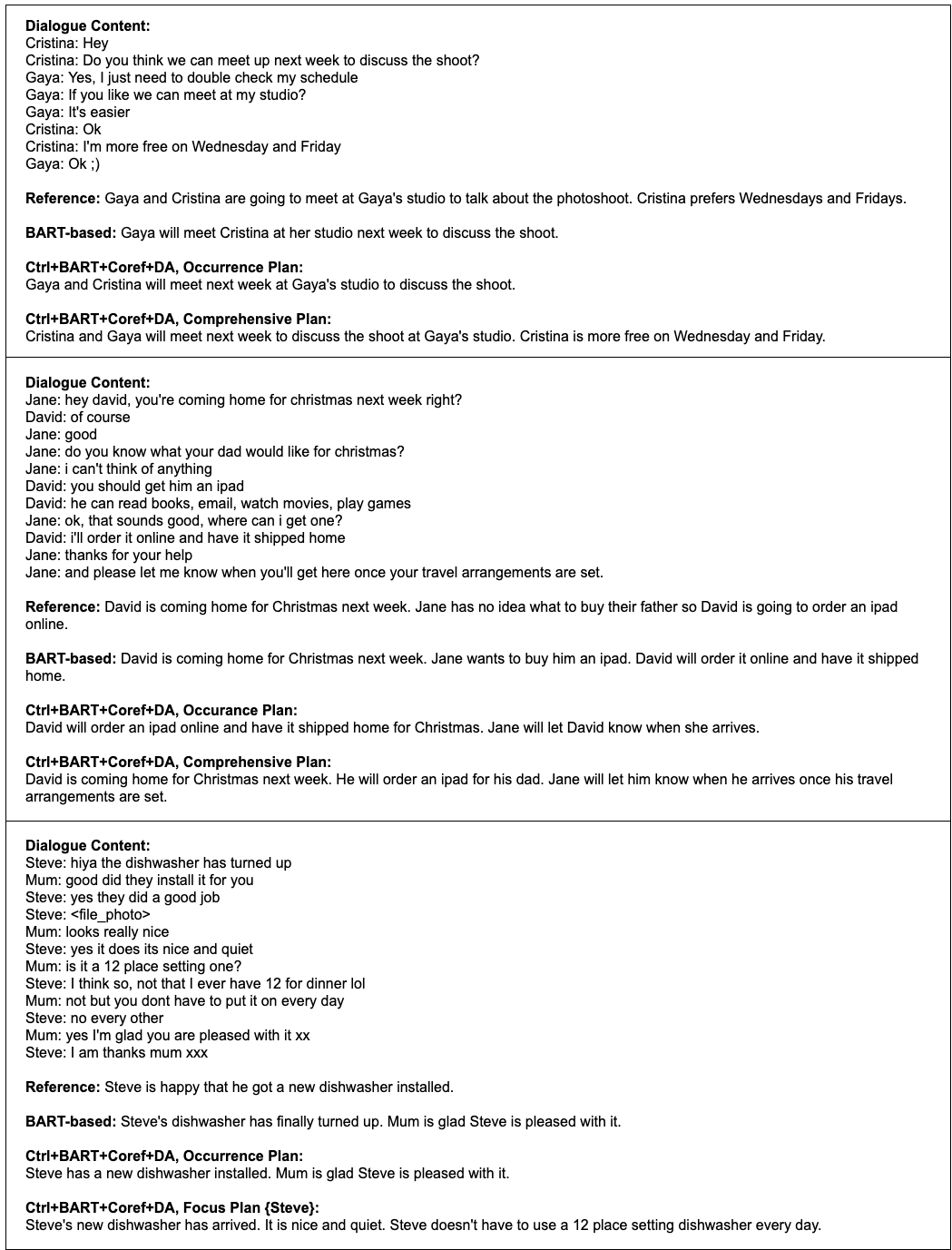}
\caption{Dialogue examples with summaries from a baseline model and our controllable generation.}
\label{sop-apx-example-fig}
\end{figure*}

\begin{figure*}[ht]
\centering
\includegraphics[width=15cm]{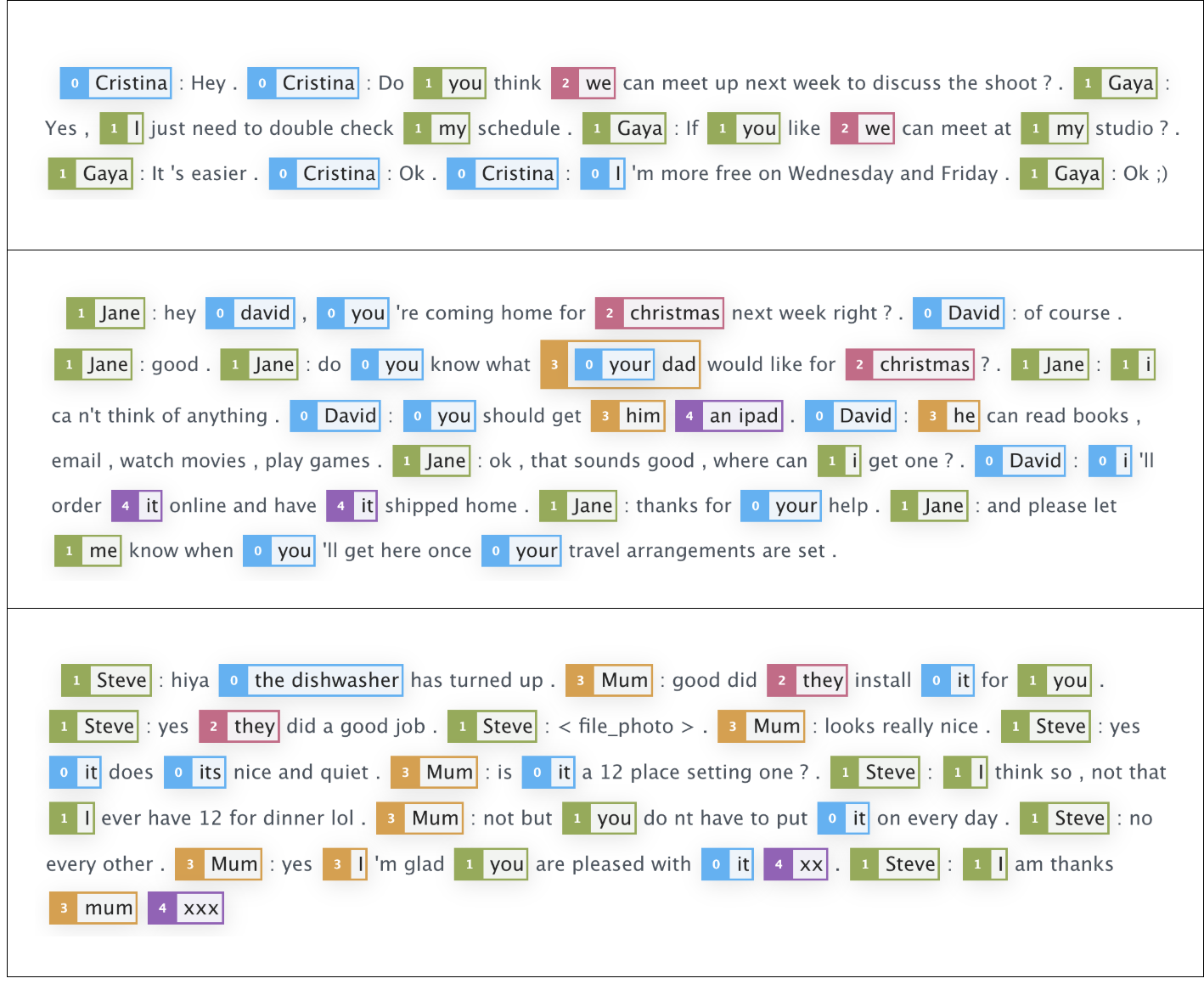}
\caption{Dialogue examples with coreference resolution information. Words/Spans in one coreference cluster are labeled with the same color. Noted that this is the original output from AllenNLP \cite{Gardner2017AllenNLP} coreference resolution tool.}
\label{sop-apx-coref-fig}
\end{figure*}

\end{document}